# *SPARCNN:* SPAtially Related Convolutional Neural Networks

JT Turner[1], Kalyan Moy Gupta[1], and David Aha[2]
[1]Knexus Research Corporation; 174 Waterfront St. Suite 310; National Harbor, MD 20745
[2]Navy Center for Applied Research in Artificial Intelligence
Naval Research Laboratory (Code 5514); Washington, DC 20375
*{first.last}*@knexusresearch.com | david.aha@nrl.navy.mil

*Abstract*— The ability to accurately detect and classify objects at varying pixel sizes in cluttered scenes is crucial to many Navy applications. However, detection performance of existing state-of-the-art approaches such as convolutional neural networks (CNNs) degrade and suffer when applied to such cluttered and multi-object detection tasks. We conjecture that spatial relationships between objects in an image could be exploited to significantly improve detection accuracy, an approach that had not yet been considered by any existing techniques (to the best of our knowledge) at the time the research was conducted. We introduce a detection and classification technique called Spatially Related Detection with Convolutional Neural Networks (SPARCNN) that learns and exploits a probabilistic representation of inter-object spatial configurations within images from training sets for more effective region proposals to use with state-of-the-art CNNs. Our empirical evaluation of SPARCNN on the VOC 2007 dataset shows that it increases classification accuracy by 8% when compared to a region proposal technique that does not exploit spatial relations. More importantly, we obtained a higher performance boost of 18.8% when task difficulty in the test set is increased by including highly obscured objects and increased image clutter.

*Index Terms*—Deep Learning, Object Detection, Convolutional Neural Networks (CNN), Keypoint Density Region Proposal (KDRP)

## I. Introduction

Applications of image processing algorithms to Navy missions such as those involving intelligence surveillance and reconnaissance (ISR), maritime security, and force protection (FP) require that they achieve high accuracy and respond in real time. Conventional approaches to image classification tasks includes the use of keypoint descriptors and local feature descriptors [1], which are binned into histograms and compared to other keypoints to match similar objects. For instance, work on deformable part models and detection of parts [1] gave rise to specialized part models that operate by transfer of likely locations [2], which achieved high classification and detection accuracy, and speed, on the fine-grained Caltech UCSD bird dataset [3]. Recently, Convolutional Neural Networks (CNNs), a deep learning approach, has emerged as a promising technique that dramatically outperforms conventional approaches on classification accuracy. Evolving from the early work of [4], which primarily focused on image classification, CNNs can now achieve state-of-the-art performance on object detection tasks [5]. Although CNNs have become adept at processing pixels to classify objects, and even computing bounding box targets based on the objectness score of the region, there is additional information about the object or objects in an image that we cannot discern from a low level pixel signal. In this paper, we present a new system for multi-object detection in images with clutter called **Spa**tially **R**elated detection with **C**onvolutional **N**eural **N**etworks (SPARCNN). SPARCNN includes the following three key features:

- It leverages and extends our previous state of the art region proposal technique called KDRP [6]; KDRP is a region proposal technique that uses density of high interest features to propose regions with higher likelihood for objects of interest.
- It recursively proposes regions based on where it has previously observed objects.
- It adjusts thresholds for object detection based on what objects have been detected with a sufficiently high confidence.

The rest of the paper is organized as follows: Section 2 presents the contributions of the SPARCNN approach to the existing detection pipeline with a subsection devoted to each of the three features enumerated above. Section 3 presents the results of SPARCNN evaluation on the VOC 2007 dataset, and Section 4 concludes the paper with a discussion and outlines our planned future work.

## II. SPARCNN

SPARCNN is designed to detect objects in a cluttered image with high accuracy. During training, two models are trained for use by SPARCNN; Fast R-CNN [5], and the **S**patial **R**elation **M**odel (SRM).

### A. SPARCNN Training

SRM is captures the following attributes about a training dataset assuming that there are *n* classes, it stores the following information:

Thanks to ONR for supporting this research.

1. *Fraction of class label*: SRM sums all objects *s* that are of a given class *a,* and creates an *n*-dimensional list of the probability of any given class.
2. *Fraction of images present*: Sums over all images *I* where there is at least one instance of an object of class *a*, and stores them in an *n*-dimensional list.
3. *Conditional Probabilities*: Given a class *a*, and another class *b*, this is calculated for a given *b* as the probability of any given image containing *a ^ b* divided by the probability of an image only containing *b*. This is stored in an *n* x *n* matrix.
4. *Spatial Probabilities*: Given a class *a*, and another class *b*, this is the normalized fraction of locations on the divided grid, as shown in Figure 1. The anchor object class *a* is defined to occupy 100% of z4 in the diagram, and the secondary object has its overlap with each other region calculated. For example, an object that is strictly above Z4 would increment (1 object * 1.00 overlap). This is done for every pair of objects in every image, and normalized, so the end result is an $n^2 * 9$ sized lookup table, where any given entry is the normalized fraction where class *b* has occurred in relation to class *a*.
5. *Relative Sizes*: Given a class *a,* and another class *b,* this is an $n^2$ x 2 dimensional list of the mean relative pixel$^2$ sizes of $\frac{a}{b}$, as well as the standard deviation of the relative sizes.
6. *Aspect Ratios*: For any arbitrary class *a*, an *n x 2* table is calculated for the mean and standard deviation of the shorter side of the image over the larger side of the image. All aspect ratios will fall in the range of (0,1].

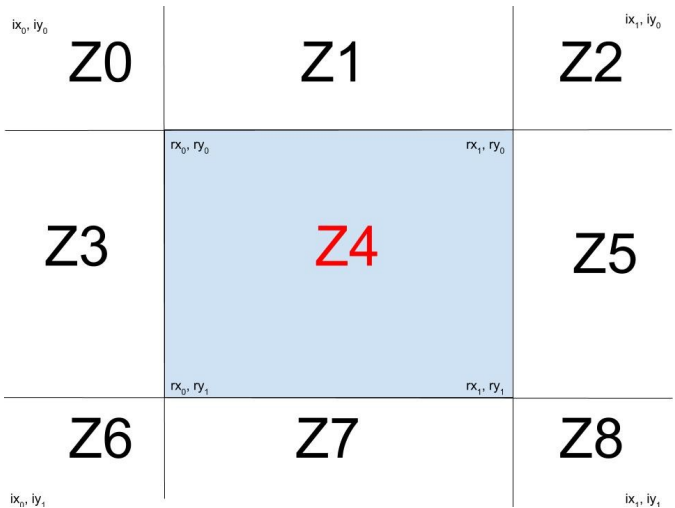

*Figure 1- Spatial Probability Locations*

Region proposal is the only difference in the training of SPARCNN versus Fast R-CNN; the SRM is also trained. The Fast R-CNN models trained in [5] can still be used to process the image corpus.

B. *Applying SPARCNN*

The application of SPARCNN varies from that of traditional Fast R-CNN, both in terms of region proposal and hypothesis selection. Both techniques leverage the SRM to search for and select objects that are overlooked by a simpler region proposal and convolution method. SPARCNN is a recursive method for object detection that proposes regions based on highly confident detections, and adjusts detection thresholds based on the objects in the image that we are confident of observing.

*1) SPARCNN Overview*

```
SPARCNN_detect(image, SRM):
    confirmed= [ ]
    for size in ['LARGE', 'MEDIUM',
'SMALL']:
        first_loop = True
        while nms_detections != None or
first_loop:
            first_loop = False
            if known_detections == None:
                regions <-- gen_kdrp_reg(
size=size, img=img)
            else:
                regions <-- gen_srm_reg(
size=size, srm=SRM, known=confirmed,
img=img)
            new_detections <--
sparcnn_detect (reg=regions, srm=SRM,
known=confirmed)
            nms_detections <-- nonmax(
reg=regions, known=confirmed)
            confirmed += nms_detections
    return confirmed
```

*Algorithm 1*: *Pseudocode for SPARCNN detection routine*

Region proposal in SPARCNN is performed in three recursive tiers based on object size; large object, medium object, and small object.

Region proposal is an iterative and recursive process. Iteration is done using three ranges of window sizes; large (where the width of the region is between 40% and 99% of the width of the image, height of the region is with 40% and 99% the height of the image), medium (constrained similarly between 10% and 64%), and small region proposal (constrained between 2% and 16%).

*2) Region Proposal*

Building on the work of KDRP [6], SPARCNN uses a keypoint density based approach for region proposal. As more objects are detected in an image, prior knowledge of co-occurring objects can be leveraged to improve the proposal of regions to search for additional objects nearby. Once KDRP has detected an object, it begins a new type of region proposal (the function *gen_srm_reg* in the pseudocode, no longer *gen_kdrp_reg*). Regions from the SRM are generated to be consistent with training set observations. **Algorithm 2** initializes *keypoints* as a blank array, and its first loop is over every detected object in the image. For each object detected with a sufficiently high probability, it loops through every class *c* observed in the training set, and counts the number of times *n* the detected object class and objects in class *c* co-occurred in the training set. Then using the spatial probability location grid of **Figure 1**, β*n* (given β=15 is a constant number to produce more keypoints and regions determined through cross

```
gen_srm_reg(size, srm, confirmed_detects, img):
  keypoints ← []
  for det in confirmed_detections:
    for class in srm.class_labels:
      co_occour ← srm.get_cooccourances(det.class, class)
      keypoints += kdrp_keypoint_gen(
                     loc=srm.loc
                     o1=det.cls,
                     o2=class)
  return kdrp_generate(size=size,
                     num_regions=β*co_occour,
                     keypoints=keypoints)
```

*Algorithm 2- SRM region generation pseudocode*

validation) keypoints are randomly generated in the corresponding sector of the grid, and their (x, y) locations are recorded.

For example, if objects of type *person* and *dog* co-occurred 73 times, SPARCNN would generate 1,095 keypoints. Suppose that 30% of dogs were located below people (box $Z_7$), 50% were found overlapping people ($Z_4$), and 5% and 15% were found to the left and right ($Z_3$ and $Z_5$ respectively). SPARCNN would mirror and split the grid on the central y axis (we assume that for everyday objects it does not matter if something is to the right or left), so $\frac{Z_3+Z_5}{2}$ = 10%, which yields new values for $Z_3$ and $Z_5$. SPARCNN would then distribute the 1,095 keypoints evenly in proportion to the spatial matrix (i.e., 329 keypoint locations under the person detection in $Z_7$, 547 keypoint locations overlapping the person, and 109 keypoints generated on the left and right side of the person, respectively). This is done for every class, and then used as input to KDRP. Usually, KDRP operates by detecting binary patterns and keypoints in the changes of gradient of the image to generate regions [6], but these keypoint locations can be given directly to KDRP so it selects high keypoint regions instead of local binary patterns [7]. Using this method, SPARCNN is likely to generate the following regions looking for a dog given information on where a person is located **(Figure 2)**.

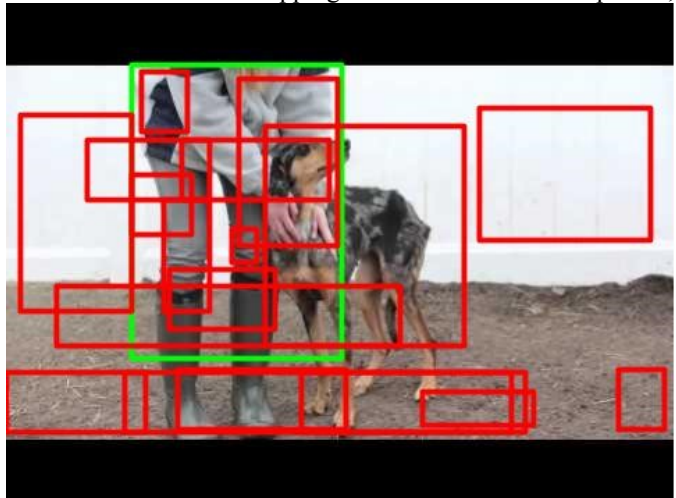

*Figure 2- Regions generated by SRM information*

*3) Hypothesis and Threshold Adjustment*

The confidence of the detections has, to this point in the SPARCNN process, been fixed; only different regions have been proposed than would have been proposed by a traditional selective search algorithm [5] or by a region proposal network [8]. Although region proposal techniques can be used to reduce time constraints [6], as long as the correct region is proposed by multiple region proposal techniques, they are unlikely to reduce detection accuracy. There are two ways to adjust hypothesis acceptance; by raising or lowering the threshold of probability needed for detection ($T_p$), or by raising or lowering the probability of the region that has been convoluted ($R_p$). $T_p$ is selected through multi fold validation to produce the maximum detection accuracy in all cases where $R_p \geq T_p$. In general, the amount that we want to change the probability is split between $R_p$, and $T_p$ such that the probability of detection will be a positive number, but a number smaller than 1.00 (since this would make it impossible to identify an object). We used cross validation to set a minimum signal strength needed for detection of .36 from the network. No matter what objects are around it, and if it matches the aspect ratio and relative size perfectly, positive detections cannot be set at lower values without spurious detections.

The three types of evidence from the SRM can be used to influence threshold or detection confidence:

1. Object aspect ratio
2. Object correlation
3. Object relative size

*a) Object Aspect Ratio*

For each object in the training set, we record the ratio of the longest side to the shortest side of the object, along with its class *c*. We do not use a fixed height and width because, for example, a bottle (which is usually a little more than twice as long as its width) could be misidentified if it were laying on its side (in a bottle rack for example). After computing these ratios, we calculate the mean aspect ratio ($A_c$) and the standard deviation of the aspect ratios ($S_c$). Because aspect ratio may be noisy (e.g., there may be an oddly shaped water bottle, or perhaps a person has a square shape due to a kneeling posture), even if the aspect ratio matches the threshold should not necessarily be changed greatly. When applied, SPARCNN computes the aspect ratio and the Z-score (number of standard deviations away from the mean), and the region probability and threshold probability are adjusted as shown in **Table 1.**

| Experimental Value x in Z score | Classification | Δ Rp | Δ Tp |
|---|---|---|---|
| -1 ≤ x ≤ 1 | Evidence for | + .02 | -.02 |
| -2 ≤x or x ≥ 2 | Neutral Evidence | 0 | 0 |
| -3 ≤x or x≥ 3 | Evidence Against | -.02 | +.02 |

*Table 1- Aspect Ratio Evidence*

*b)*   *Object Correlation*

The increase in SPARCNN's accuracy is primarily due to the use of object correlations to boost detections. SPARCNN creates a copy of the image, but instead of 3 pixel values at each *(x, y)* coordinate, it assigns a probability modifier for each class. After an object in class *A* is detected, then for every object in class *B,* SPARCNN, will update every pixel using the probability modifier to reflect changes in probability of all the classes. In the SRM, let the fraction of objects that occurred in the same spatial position with respect to *A* be $S_A$, and let $P_B = P(B|A)$. SPARCNN modifies the value needed for detection as follows: $T_p = T_p - (S_A * P_B)$. This ensures that objects that are conditionally codependent will lower the threshold, and the effect is even greater if they were in a previously detected spatial relation. Each pixel on the representation of an image is assigned a new threshold weight. To determine the threshold needed for any given region, SPARCNN sums all of the pixel value thresholds contained in that region, and averages them.

*c)*   *Relative Object Size*

For every pair of objects in the training images, the relative size of every object is recorded. The means and standard deviations of the class wise pairs are computed. Much like aspect ratio, this is a weak evidence for object identification, as objects that are near or far from the camera may appear to be incorrect in object size, but actually be a real detection, as highlighted in **Figure 3.** This is considered weaker evidence than aspect ratio.

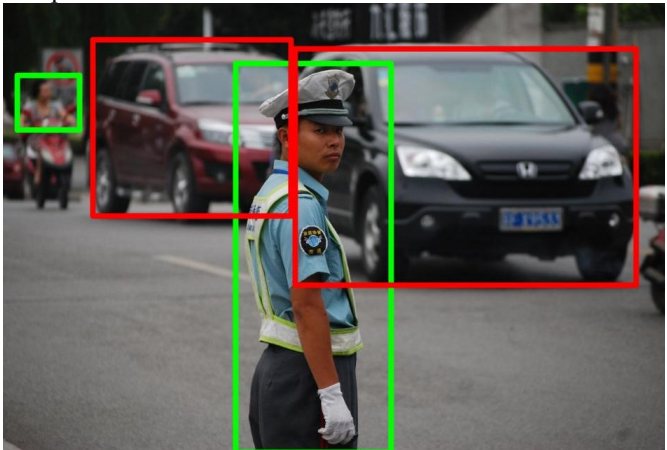

*Figure 3- Two people are visible; one is larger than the cars and one is much smaller*

| Experimental Value x in Z score | Classification | Δ Rp | Δ Tp |
|---|---|---|---|
| -1 ≤ x ≤ 1 | Evidence for | + .02 | -.01 |
| Else | Inconclusive | 0 | 0 |

*Table 2- Relative Object Size Evidence*

## III. EXPERIMENTS

*A. Objective & Hypotheses*

*Our overall objective is to assess whether, by leveraging (1) spatial relationships between objects and (2) conditional probabilities as described in Section 3, SPARCNN would outperform neural networks using the same region proposal techniques and network topology.*

Our first hypothesis ($H_1$) is that by adding additional objects, the increase in recall from previously overlooked objects will be greater than the false positives that arise from misidentifying objects, so we expect an increase in accuracy and $F_1$ measure

$$H_1: \text{Accuracy}_{(SPARCNN)} > \text{Accuracy}_{(BASELINE)}$$

Our second hypothesis ($H_2$) is that even with the now spurious false positives from SPARCNN, the added true positives will increase accuracy and $F_1$ measure enough such that the Area under the ROC curve (AUC) will be no less than the AUC of the baseline. Stated formally:

$$H_2: \text{AUC}_{(SPARCNN)} = \text{AUC}_{(BASELINE)}$$

We test $H_1$ using an A/B Split test, and $H_2$ using a class-wise paired t test. We used the very deep network full model VGG-16 [9] trained using Fast R-CNN [5]. We set hyper parameters and SRM evidence levels (Tables 1 and 2) using 5-fold cross validation on a held-out data set. In this experiment, we compared two systems.

Baseline: uses KDRP + Fast R-CNN without using SRM for region proposal or hypothesis selection.
SPARCNN: uses the additional region proposals and hypothesis selection criteria, and undergoes hypothesis changes and detection threshold adjustment as described in Section 2.

*B. Datasets*

We tested SPARCNN only with PASCAL VOC 2007. The dataset split and annotations were the same as used in [5], and dataset characteristics are given in **Table 3.**

| Characteristic | Value |
|---|---|
| Number of classes | 20 |
| Class Distribution | Skewed (Minimum class "dining table" has 359 training instances, maximum class "people" has 7,957 instances) |
| Objects per image | 1-42 |
| Target object size (pixel$^2$) | 44 – 248,003 |
| Train/Test split | 8539/1424 |

*Table 3- PASCAL VOC 2007 characteristics.*

PASCAL VOC 2007 also has a difficult flag that can be toggled *True* or *False*. An object in the image can be labeled as "difficult" for several reasons, most often because it is cropped or mostly not shown in the image, as exemplified in **Figure 4.**

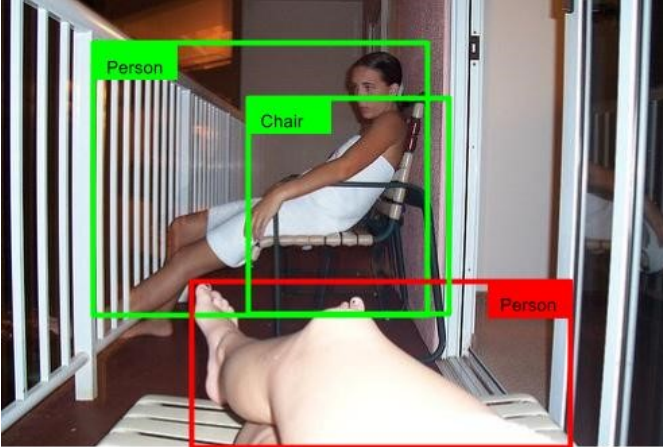

*Figure 4- The two green objects are not difficult because they are entirely visible, but the person who we can only see the legs of is considered difficult.*

### C. Evaluation Metrics

We measured the algorithm using the following measures: accuracy, recall, precision, the $F_1$ measure, and AUC. Three outcomes were recorded for each detection attempt/undetected object:

- *True Positive (TP)*: A true positive is recorded if the predicted bounding box has an intersection over union (IoU) or area greater than 0.5, and is of the correct class.
- *False Positive (FP)*: A false positive is recorded for every detection that does not have an IoU of greater than 0.5 with a previously undetected object of the correct class.
- *False Negative (FN)*: A false negative is recorded if none of the system detections match the ground truth bounding box for IoU and class label.

Using these definitions, we define the following four terms:

- $Accuracy = \frac{TP}{TP+FP+FN}$
- $Recall = \frac{TP}{TP+FN}$
- $Precision = \frac{TP}{TP+FP}$
- $F_1\ Measure = \frac{2*TP}{(2*TP)+FP+FN}$

We calculate the AUC as the interpolated mAP, as described in [10].

### D. Results

Tables 4 and 5 show the results (for the first four metrics) for two dataset conditions: (1) without and (2) with difficult annotations, where the boldfaced number indicates the system that significantly performed better.

| METRIC | BASELINE | SPARCNN | %CHANGE |
|---|---|---|---|
| ACCUR % | 45.95 | **49.29** | 7.27 |
| RECALL % | 51.72 | **66.78** | 29.12 |
| PRECIS % | **80.48** | 65.3 | -16.86 |
| $F_1$ MEAS % | 62.97 | **66.04** | 4.88 |

*Table 4- PASCAL VOC 2007 Evaluation w/o difficult annotations*

| METRIC | BASELINE | SPARCNN | %CHANGE |
|---|---|---|---|
| ACCUR % | 39.89 | **47.37** | 18.75 |
| RECALL % | 42.92 | **58.17** | 35.53 |
| PRECIS % | **84.97** | 71.84 | -15.45 |
| $F_1$ MEAS % | 57.04 | **64.29** | 12.71 |

*Table 5- PASCAL VOC 2007 Evaluation with difficult annotations*

For both datasets, SPARCNN outperformed Baseline on accuracy, recall, and $F_1$, but performed worse on precision. This is because SPARCNN adds detections that would have been skipped due to lower confidence than the needed threshold. Although SPARCNN does this correctly more often than not (as evidenced by the higher accuracy and $F_1$ measure), it also creates additional false positives, which reduces precision. The A/B split testing for both the standard and difficult splits are statistically significant at a level of α=.05, so we accept the hypothesis $H_1$.

We also found that SPARCNN increases relative performance for difficult (i.e., cluttered, overlapping) scenes. The percentage change from the non-difficult to difficult dataset conditions, in comparison with Baseline, is more than double, and the $F_1$ measure increases three-fold, while the percentage change in precision actually decreases. As more clutter and obfuscation of ground truth target objects are added to an image, fewer false positives result, which increases precision.

In the more commonly used metric for PASCAL VOC 2007 evaluation (AUC), there was no significant difference at a level of α=0.05 between the baseline (0.6474) and SPARCNN (0.6431). A classwise comparison is shown in **Figure 5.**

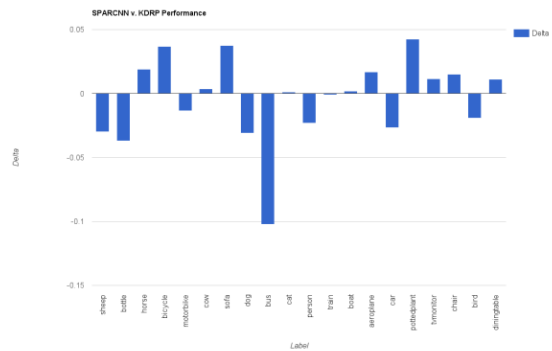

*Figure 5- Classwise comparison of AUC for SPARCNN v. Baseline*

## IV. CONCLUSION AND DISCUSSION

Although SPARCNN did not outperform Baseline for AUC, this metric does not accurately highlight its improvements. Using the same cross validation scheme to select parameters for SPARCNN to use for detection threshold levels as the baseline algorithm, there exists a set of parameter settings for which SPARCNN significantly outperforms Baseline in terms of object recall, while also increasing accuracy and $F_1$.

This study warrants future work in possible improvements to SPARCNN so that it can be applied in real-time tasks that require instantaneous monitoring and detection. For example, we plan to use a different network topology that would propose regions automatically as part of convolution as seen in [8], using information about object semantics and what can physically exist, and using different trained networks on different sized objects for our large, medium, and small search regions.